\documentclass[10pt,twocolumn,letterpaper]{article}

\usepackage{wacv}
\usepackage{times}
\usepackage{epsfig}
\usepackage{graphicx}
\usepackage{amsmath}
\usepackage{amssymb}

\wacvfinalcopy 

\def\wacvPaperID{638} 

\ifwacvfinal\fi
\begin{document}

\title{Scale Match for Tiny Person Detection}

\author{
Xuehui Yu \hspace{1cm} Yuqi Gong \hspace{1cm} Nan Jiang\hspace{1cm} Qixiang Ye \hspace{1cm} Zhenjun Han\thanks{corresponding author} \\
University of Chinese Academy of Sciences, Beijing, China\\
{\tt\small \{yuxuehui17,gongyuqi18,jiangnan18\}@mails.ucas.ac.cn\hspace{1cm}\{qxye,hanzhj\}@ucas.ac.cn}
}

\maketitle
\ifwacvfinal\thispagestyle{empty}\fi

\begin{abstract}
 Visual object detection has achieved unprecedented advance with the rise of deep convolutional neural networks. However, detecting tiny objects (for example tiny persons less than 20 pixels) in large-scale images remains not well investigated. The extremely small objects raise a grand challenge about feature representation while the massive and complex backgrounds aggregate the risk of false alarms. In this paper, we introduce a new benchmark, referred to as TinyPerson, opening up a promising direction for tiny object detection in a long distance and with massive backgrounds. We experimentally find that the scale mismatch between the dataset for network pre-training and the dataset for detector learning could deteriorate the feature representation and the detectors. Accordingly, we propose a simple yet effective Scale Match approach to align the object scales between the two datasets for favorable tiny-object representation. Experiments show the significant performance gain of our proposed approach over state-of-the-art detectors, and the challenging aspects of TinyPerson related to real-world scenarios. The TinyPerson benchmark and the code for our approach will be publicly available\footnote{https://github.com/ucas-vg/TinyBenchmark}.
\end{abstract}

\begin{figure}
\begin{center}
    \begin{tabular}{ccc}
    \includegraphics[width=8.3cm]{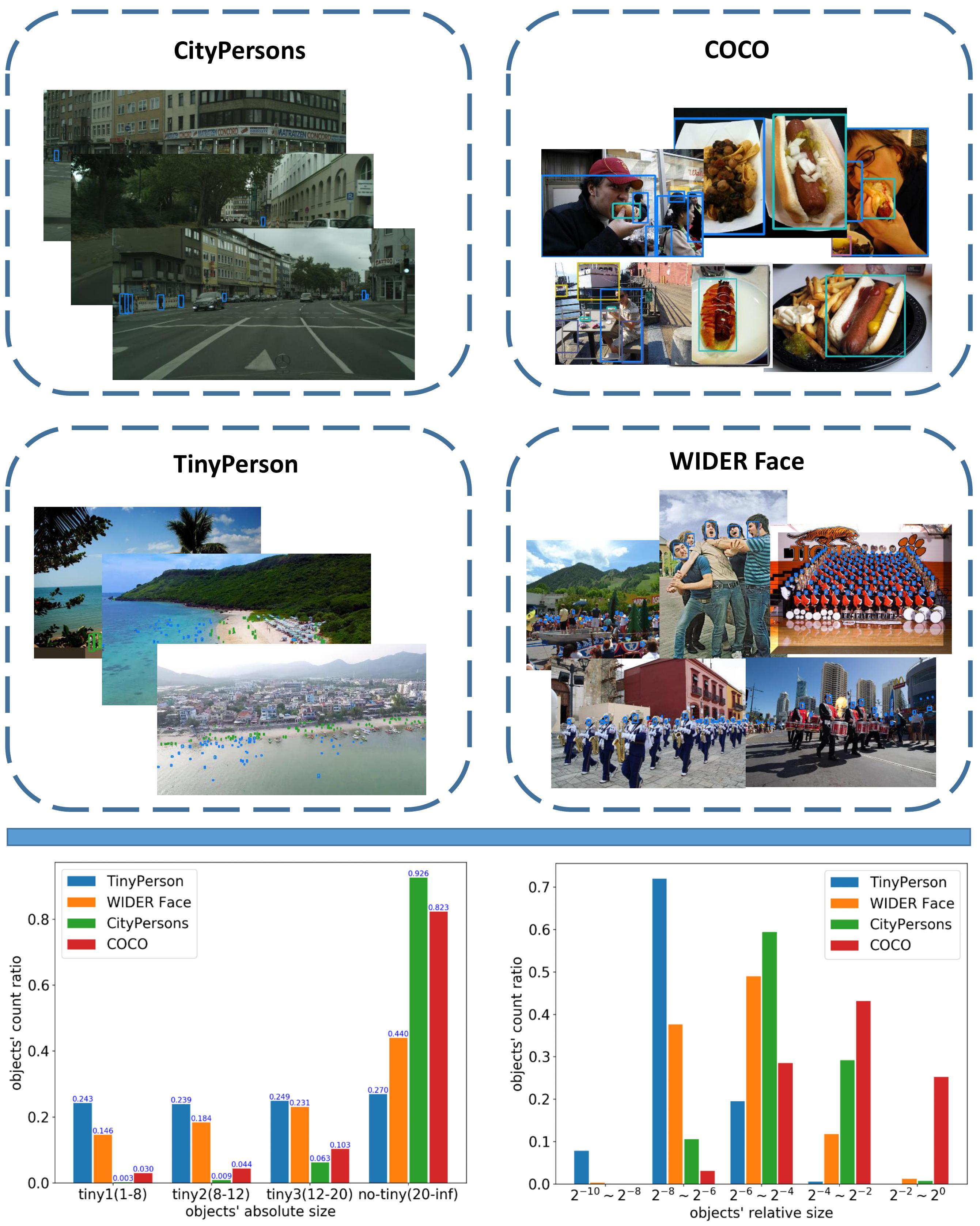}
    \end{tabular}
	\caption{Comparison of TinyPerson with other datasets. \textbf{Top}: Image examples from TinyPerson show the great challenges (Please zoom in for details).
	\textbf{Bottom}: Statistics about absolute size and relative size of objects.
	}
\label{fig:motivation}
\end{center}
\end{figure}

\section{Introduction}
\label{sec:intro}
\noindent Person/pedestrian detection is an important topic in the computer vision community  \cite{enzweiler2008monocular} \cite{dollar2011pedestrian} \cite{geiger2012we} \cite{zhang2017citypersons} \cite{mao2017can} \cite{zhang2017towards}, with wide applications including surveillance, driving assistance, mobile robotics, and maritime quick rescue. 
With the rise of deep convolutional neural networks, pedestrian detection has achieved an unprecedented progress. Nevertheless, detecting tiny persons remains far from well explored. 

The reason about the delay of the tiny-person detection research is lack of significant benchmarks. The scenarios of existing person/pedestrian benchmarks \cite{dalal2005histograms}\cite{ess2008mobile}\cite{wojek2009multi}\cite{enzweiler2008monocular}\cite{dollar2011pedestrian}\cite{geiger2012we}， $e.g.,$ CityPersons \cite{zhang2017citypersons}, are mainly in a near or middle distance. They are not applicable to the scenarios where persons are in a large area and in a very long distance, $e.g.$, marine search and rescue on a helicopter platform.



Different from objects in proper scales, the tiny objects are much more challenging due to the extreme small object size and low signal noise ratio, as shown in Figure \ref{fig:motivation}. After the video encoding/decoding procedure, the image blur causes the tiny objects mixed with the backgrounds, which makes it require great human efforts when preparing the benchmark. 
The low signal noise ratio can seriously deteriorate the feature representation and thereby challenges the state-of-the-art object detectors.

%

To detect the tiny persons, we propose a simple yet effective approach, named Scale Match. The intuition of our approach is to align the object scales of the dataset for pre-training and the one for detector training.
The nature behind Scale Match is that it can better investigate and utilize the information in tiny scale, and make the convolutional neural networks (CNNs) more sophisticated for tiny object representation. The main contributions of our work include:

1. We introduce TinyPerson, under the background of maritime quick rescue, and raise a grand challenge about tiny object detection in the wild. To our best knowledge, this is the first benchmark for person detection in a long distance and with massive backgrounds. The train/val. annotations will be made publicly and an online benchmark will be setup for algorithm evaluation.

2. We comprehensively analyze the challenges about tiny persons and propose the Scale Match approach, with the purpose of aligning the feature distribution between the dataset for network pre-training and the dataset for detector learning. 

3. The proposed Scale Match approach improves the detection performance over the state-of-the-art detector (FPN) with a significant margin (~5\%).  


\section{Related Work}
\noindent{\bf Dataset for person detection:} Pedestrian detection has always been a hot issue in computer vision. Larger capacity, richer scenes and better annotated pedestrian datasets,such as INRIA \cite{dalal2005histograms}, ETH \cite{ess2008mobile}, TudBrussels \cite{wojek2009multi}, Daimler \cite{enzweiler2008monocular}, Caltech-USA \cite{dollar2011pedestrian}, KITTI \cite{geiger2012we} and CityPersons \cite{zhang2017citypersons} represent the pursuit of more robust algorithms and better datasets. The data in some datasets were collected in city scenes and sampled from annotated frames of video sequences. Despite the pedestrians in those datasets are in a relatively high resolution and the size of the pedestrians is large, this situation is not suitable for tiny object detection. 

TinyPerson represents the person in a quite low resolution, mainly less than 20 pixles, in maritime and beach scenes. Such diversity enables models trained on TinyPerson to well generalize to more scenes, $e.g.,$ Long-distance human target detection and then rescue.

Several small target datasets including WiderFace \cite{yang2016wider} and TinyNet \cite{pang2019r2} have been reported. TinyNet involves remote sensing target detection in a long distance. However, the dataset is not publicly available. WiderFace mainly focused on face detection, as shown in Figure \ref{fig:motivation}. The faces hold a similar distribution of absolute size with the TinyPerson, but have a higher resolution and larger relative sizes, as shown in Figure \ref{fig:motivation}.

\noindent{\bf CNN-based person detection:} In recent years, with the development of Convolutional neural networks (CNNs), the performance of classification, detection and segmentation on some classical datasets, such as ImageNet \cite{deng2009imagenet}, Pascal \cite{everingham2010pascal}, MS COCO \cite{lin2014microsoft}, has far exceeded that of traditional machine learning algorithms.Region convolutional neural network (R-CNN) \cite{girshick2014rich} has become the popular detection architecture. OverFeat adopted a Conv-Net as a sliding window detector on an image pyramid. R-CNN adopted a region proposal-based method based on selective search and then used a Conv-Net to classify the scale normalized proposals. Spatial pyramid pooling (SPP) \cite{he2015spatial} adopted R-CNN on feature maps extracted on a single image scale, which demonstrated that such region-based detectors could be applied much more efficiently. Fast R-CNN \cite{girshick2015fast} and Faster R-CNN \cite{ren2015faster} made a unified object detector in a multitask manner. Dai et al. \cite{dai2016r} proposed R-FCN, which uses position-sensitive RoI pooling to get a faster and better detector. 

While the region-based methods are complex and time-consuming, single-stage detectors, such as YOLO \cite{redmon2016you} and SSD \cite{liu2016ssd}, are proposed to accelerate the processing speed but with a performance drop, especially in tiny objects. 

\noindent{\bf Tiny object detection:} 
Along with the rapid development of CNNs, 
researchers search frameworks for tiny object detection specifically. Lin et al. \cite{lin2017feature} proposed feature pyramid networks that use the top-down architecture with lateral connections as an elegant multi-scale feature warping method. Zhang et al. \cite{zhang2017s3fd} proposed a scale-equitable face detection framework to handle different scales of faces well. Then J Li et al. \cite{li2019dsfd} proposed DSFD for face detection, which is SOTA open-source face detector. Hu et al. \cite{hu2017finding} showed that the context is crucial and defines the templates that make use of massively large receptive fields. Zhao et al. \cite{zhao2017pyramid} proposed a pyramid scene-parsing network that employs the context reasonable. Shrivastava et al. \cite{shrivastava2016training} proposed an online hard example mining method that can improve the performance of small objects significantly. 
\section{Tiny Person Benchmark}
\noindent In this paper, the size of object is defined as the square root of the object's bounding box area. We use $G_{ij}=(x_{ij}, y_{ij}, w_{ij}, h_{ij})$ to describe the $j$-th object's bounding box of $i$-th image $I_i$ in dataset, where $(x_{ij}, y_{ij})$ denotes the coordinate of the left-top point, and $w_{ij}, h_{ij}$ are the width and height of the bounding box. $W_i$, $H_i$ denote the width and height of $I_i$, respectively. Then the absolute size and relative size of a object are calculated as: 

\begin{small}\begin{equation}AS(G_{ij})=\sqrt{w_{ij} * h_{ij}}.\end{equation}\end{small}
\begin{small}\begin{equation}RS(G_{ij})=\sqrt{\frac{w_{ij}*h_{ij}}{W_i*H_i}}.\end{equation}\end{small}

For the size of objects we mentioned in the following, we use the objects' absolute size by default.

\subsection{Benchmark description}
\noindent\textbf{Dataset Collection:} The images in TinyPerson are collected from Internet. Firstly, videos with a high resolution are collected from different websites. Secondly, we sample images from video every 50 frames. Then we delete images with a certain repetition (homogeneity). We annotate 72651 objects with bounding boxes by hand. 
\newline
\textbf{Dataset Properties:} 
1) The persons in TinyPerson are quite \textbf{tiny} compared with other representative datasets, shown in Figure \ref{fig:motivation} and Table \ref{tab:mean and std}, which is the main characteristics of TinyPerson; 2) The \textbf{aspect ratio} of persons in TinyPerson has a large variance, given in Talbe \ref{tab:mean and std}. Since the various poses and viewpoints of persons in TinyPerson, it brings more complex diversity of the persons, and leads to the detection more difficult. In addition, TinyPerson can also make a effective supplement to the existing datasets in the diversity of poses and views aspect; 3) In TinyPerson, we mainly focus on person around seaside, which can be used for \textbf{quick maritime rescue and defense around sea}; 4) There are many images with \textbf{dense} objects (more than 200 persons per image) in TinyPerson. Therefore, the TinyPerson also can be used for other tasks, e.g. person counting. 

\begin{table}
\small
    \begin{center}
    \begin{tabular}{|l|c|c|c|}
        \hline
        dataset & absolute size  & relative size & aspect ratio \\
        \hline\hline
        TinyPerson        & 18.0$\pm$17.4  & 0.012$\pm$0.010 & 0.676$\pm$0.416 \\
        COCO              & 99.5$\pm$107.5 & 0.190$\pm$0.203 & 1.214$\pm$1.339 \\
        Wider face        & 32.8$\pm$52.7  & 0.036$\pm$0.052 & 0.801$\pm$0.168 \\
        CityPersons       & 79.8$\pm$67.5  & 0.055$\pm$0.046 & 0.410$\pm$0.008 \\
        \hline
    \end{tabular}
    \end{center}
    \caption{Mean and standard deviation of absolute size, relative size and aspect ratio of the datasets: TinyPerson, MS COCO, Wider Face and CityPersons.}
    \label{tab:mean and std}
\end{table}

\begin{figure}
\begin{center}
    \begin{tabular}{ccc}
    \includegraphics[width=8.3cm]{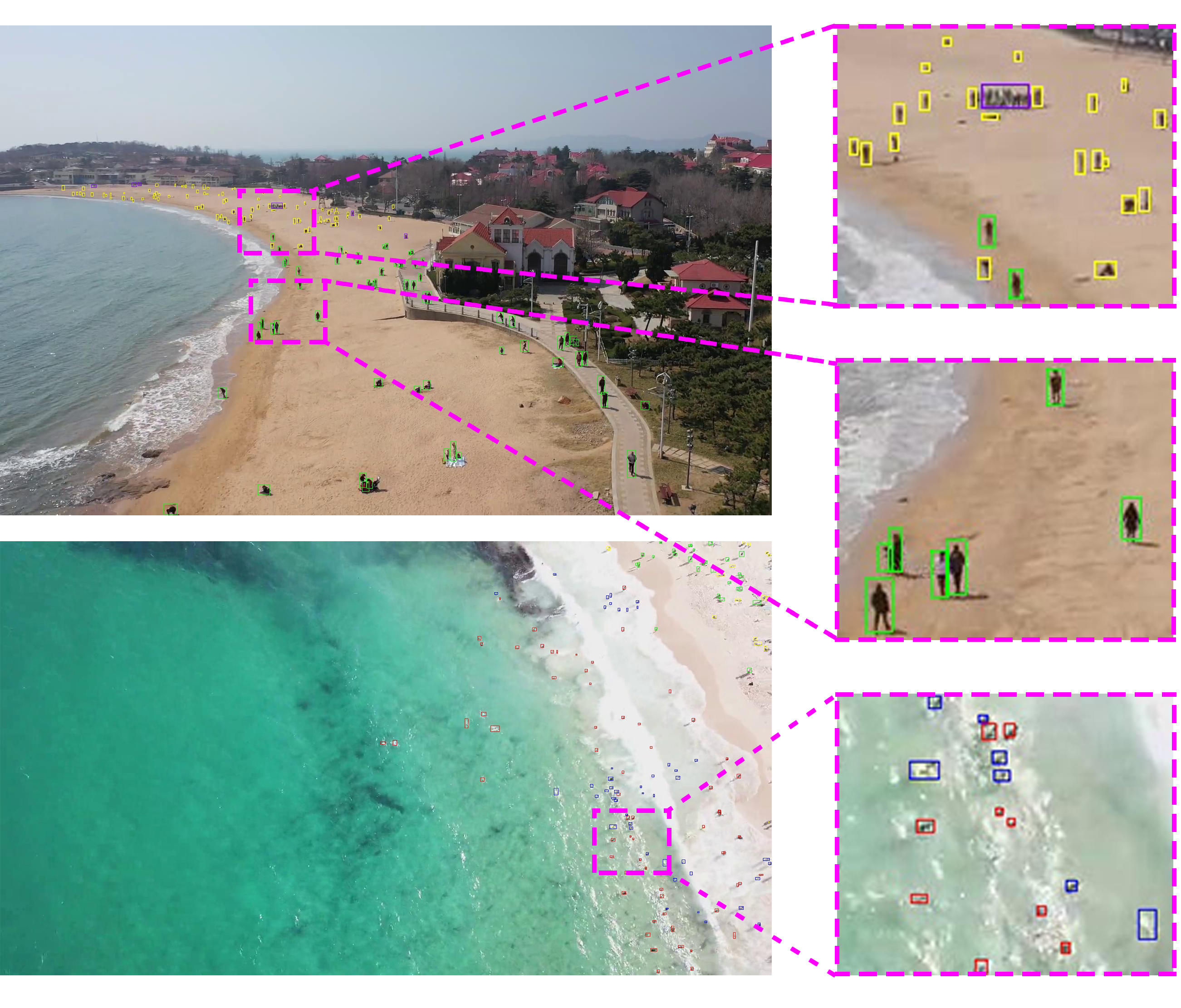}
    \end{tabular}
   \caption{The annotation examples. ``sea person'', ``earth person'', ``uncertain sea person'', ``uncertain earth person'', ignore region are represented with red, green, blue, yellow, purple rectangle, respectively. 
   The regions are zoomed in and shown on right.}
\label{fig:annotation}
\end{center}
\end{figure}

\noindent\textbf{Annotation rules:} In TinyPerson, we classify persons as ``sea person'' (persons in the sea) or ``earth person'' (persons on the land). We define four rules to determine which the label a person belongs to: 1) Persons on boat are treated as ``sea person''; 2) Persons lying in the water are treated as ``sea person''; 3) Persons with more than half body in water are treated as ``sea person''; 4) others are treated as ``earth person''. In TinyPerson, there are three conditions where persons are labeled as ``ignore'': 1) Crowds, which we can recognize as persons. But the crowds are hard to separate one by one when labeled with standard rectangles; 2) Ambiguous regions, which are hard to clearly distinguish whether there is one or more persons, and 3) Reflections in Water. In TinyPerson, some objects are hard to be recognized as human beings, we directly labeled them as ``uncertain''. Some annotation examples are given in Figure \ref{fig:annotation}.

\begin{table}
\small
    \centering
    \begin{tabular}{|l|c|c|c|}
        \hline
        TinyPerson & Train set  & valid set & sum \\
        \hline \hline
        \#image       & 794 & 816 & 1610 \\
        \#annotations & 42197 & 30454 & 72651 \\
        \hline \hline
        \#normal      & 18433 & 13787 & 32120 \\
        \#ignore      & 3369  & 1989  & 5358 \\
        \#uncertain   & 3486	 & 2832	 & 6318\\
        \#dense       &16909	 &11946  & 28855\\
        \hline \hline
        \#sea         & 26331 & 15925 & 42256\\
        \#earth       & 15867 & 14530 & 30397 \\
        \#ignore      &3369	 & 1989	 & 5358\\
        \hline
    \end{tabular}
    \caption{Statistic information in details for TinyPerson. The TinyPerson can be divided into ``normal'', ``ignore'', ``uncertain'', ``dense'' based on the attributes and ``sea'', ``earth'', ``ignore'' by the classes, which is described as annotation rules in section 3.1.}
    \label{tab:statistic information in TinyPerson.}
\end{table}

\noindent\textbf{Evaluation:} We use both AP (average precision) and MR (miss rate) for performance evaluation. For more detailed experimental comparisons, the size range is divided into 3 intervals: tiny[2, 20], small[20, 32] and all[2, inf]. And for tiny[2, 20], it is partitioned into 3 sub-intervals: tiny1[2, 8], tiny2[8, 12], tiny3[12, 20]. And the IOU threshold is set to 0.5 for performance evaluation. Due to many applications of tiny person detection concerning more about finding persons than locating precisely ($e.g.,$ shipwreck search and rescue), the IOU threshold 0.25 is also used for evaluation. 

\begin{figure}
\begin{center}
    \begin{tabular}{ccc}
    \includegraphics[width=7cm]{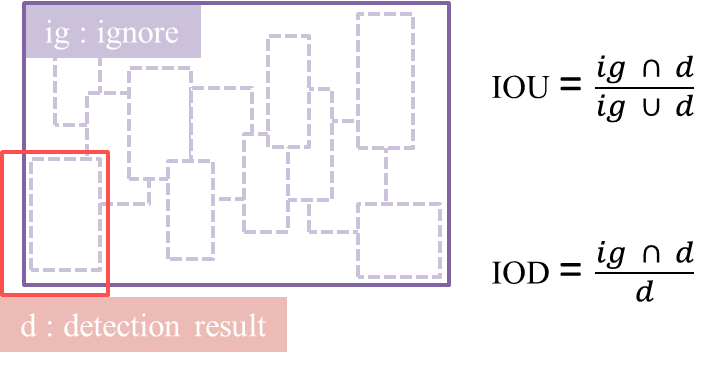}
    \end{tabular}
   \caption{IOU (insertion of union) and IOD (insertion of detection). IOD is for ignored regions for evaluation. The outline (in violet) box represents a labeled ignored region and the dash boxes are unlabeled and ignored persons. The red box is a detection's result box that has high IOU with one of ignored person.}
\label{fig:IOU_IOD}
\end{center}
\end{figure}

For Caltech or CityPersons, IOU criteria is adopted for performance evaluation. The size of most of Ignore region in Caltech and CityPersons are same as that of a pedestrian. However in TinyPerson, most of ignore regions are much larger than that of a person. Therefore, we change IOU criteria to IOD for ignore regions (IOD criteria only applies to ignore region, for other classes still use IOU criteria),as shown in Figure \ref{fig:IOU_IOD}. In this paper, we also treat uncertain same as ignore while training and testing.

\noindent\textbf{Training\&Test Set:} The training and test sets are constructed by randomly splitting the images equally into two subsets, while images from same video can not split to same subset.

Focusing on the person detection task, we treat ``sea person'' and ``earth person'' as one same class (“person”). And for detection task, we only use these images which have less than 200 valid persons. What's more, the TinyPerson can be used for more tasks as motioned before based on the different configuration of the TinyPerson manually. 
\newline

\subsection{Dataset Challenges}
\noindent\textbf{Tiny absolute size:} For a tiny object dataset, extreme small size is one of the key characteristics and one of the main challenges. To quantify the effect of absolute size reduction on performance, we down-sample CityPersons by 4*4 to construct \textbf{tiny CityPersons}, where mean of objects' absolute size is same as that of TinyPerson. Then we train a detector for CityPersons and tiny Citypersons, respectively, the performance is shown in Table \ref{tab:tiny hard}. The performance drops significantly while the object's size becomes tiny. In Table \ref{tab:tiny hard}, the $MR_{50}^{tiny}$ of tiny CityPersons is 40\% lower than that of CityPersons. Tiny objects' size really brings a great challenge in detection, which is also the main concern in this paper.

The FPN pre-trained with MS COCO can learn more about the objects with the representative size in MS COCO, however, it is not sophisticated with the object in tiny size. The big difference of the size distribution brings in a significant reduction in performance. In addition, as for tiny object, it will become blurry, resulting in the poor semantic information of the object. The performance of deep neural network is further greatly affected. 

\noindent\textbf{Tiny relative size:} 
Although tiny CityPersons holds the similar absolute size with TinyPerson. Due to the whole image reduction, the relative size keeps no change when down-sampling. Different from tiny CityPersons, the images in TinyPerson are captured far away in the real scene. The objects' relative size of TinyPerson is smaller than that of CityPersons as shown in bottom-right of the Figure \ref{fig:motivation}. 
\begin{table}
\small
    \centering
    \begin{tabular}{|l|c|c|c|c|}
        \hline
        dataset & $MR^{tiny}_{50}$ & $AP^{tiny}_{50}$\\
        \hline \hline
        tiny Citypersons       & 75.44 & 19.08\\
        3*3 tiny Citypersons   & 45.49 & 35.39\\
        \hline \hline
        TinyPerson             & 85.71 & 47.29 \\
        3*3 TinyPerson         & 83.21 & 52.47\\
        \hline
    \end{tabular}
    \caption{The performance of the tiny CityPersons, TinyPerson and their 3*3 up-sampled datasets (Due to out of memory caused by the 4*4 upsampling strategy for TinyPerson, here we just use the 3*3 up-sampling strategy as an alternative).}
\label{tab:up sample}
\end{table}

To better quantify the effect of the tiny relative size, we obtain two new datasets 3*3 tiny CityPersons and 3*3 TinyPerson by directly 3*3 up-sampling tiny CityPersons and TinyPerson, respectively. Then FPN detectors are trained for 3*3 tiny CityPersons and 3*3 TinyPerson. The performance results are shown in table \ref{tab:up sample}. For tiny CityPersons, simply up-sampling improved MR$^{tiny}_{50}$ and AP$^{tiny}_{50}$ by 29.95 and 16.31 points respectively, which are closer to the original CityPersons's performance. However, for TinyPerson, the same up-sampling strategy obtains limited performance improvement. The tiny relative size results in more false positives and serious imbalance of positive/negative, due to massive and complex backgrounds are introduced in a real scenario. The tiny relative size also greatly challenges the detection task.
\newline

\section{Tiny Person Detection}
\begin{figure*}
\begin{center}
    \begin{tabular}{ccc}
    \includegraphics[width=14.5cm]{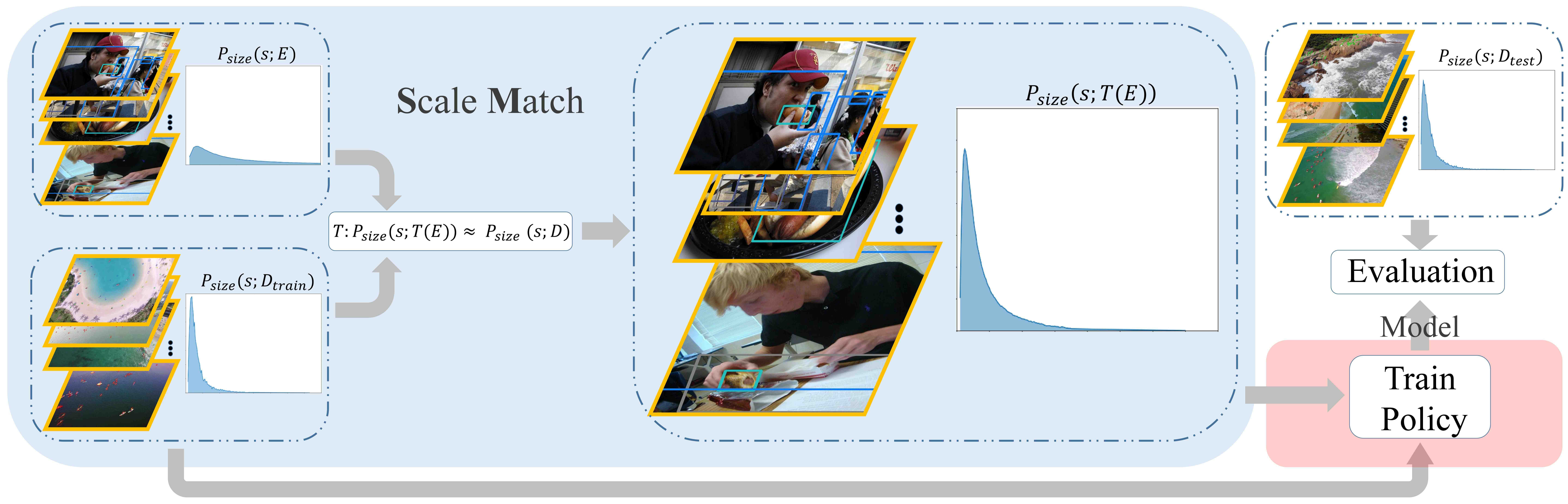}
    \end{tabular}
    \caption{The framework of Scale Match for detection. With the distributions of $E$ and $D_{train}$ dataset, the proposed Scale Match $T$($\cdot$) is adopted to adjust the $P_{size}(s;E)$ to $P_{size}(s;D_{train}$). Various training policy can be used here, such as joint training or pre-training.}
\label{fig:scale match}
\end{center}
\end{figure*}


\setlength{\tabcolsep}{12pt}\begin{table*}
\small
    \centering
    \begin{tabular}{|l|c|c|c|c|c|}
        \hline
        dataset & $MR^{tiny1}_{50}$ & $MR^{tiny2}_{50}$ &  $MR^{tiny3}_{50}$ & $MR^{tiny}_{50}$ & $MR^{small}_{50}$\\
        \hline \hline
        CityPersons   &56.40	&24.29&	8.93& 35.65 &	7.43\\
        tiny CityPersons  & 94.04&72.56	&49.37&	75.44 & 23.70 \\
        \hline
    \end{tabular}
    \caption{The performance of CityPersons and tiny CityPersons. To guarantee the objectivity and fairness,  $MR_{50}^{tiny}$, $MR_{50}^{tiny1}$,  $MR_{50}^{tiny2}$, $MR_{50}^{tiny3}$,  $MR_{50}^{small}$ are calculated with size in range [2, 20], [2, 8], [8, 12], [12, 20], [20, 32] for tiny CityPersons and in range [8, 80], [8, 32]  [32, 48], [48, 80], [80, 128] for CityPersons, respectively.}
\label{tab:tiny hard}
\end{table*}

\noindent It is known that the more data used for training, the better performance will be. However, the cost of collecting data for a specified task is very high. A commonly approah is training a model on the extra datasets as pre-trained model, and then fine-tune it on a task-specified dataset. Due to the huge data volume of these datasets, the pre-trained model sometimes boost the performance to some extent. However, the performance improvement is limited, when the domain of these extra datasets differs greatly from that of the task-specified dataset. How can we use extra public datasets with lots of data to help training model for specified tasks, $e.g.,$ tiny person detection? 

The publicly available datasets are quite different from TinyPerson in object type and scale distribution, as shown in Figure \ref{fig:motivation}. Inspired by the Human Cognitive Process that human will be sophisticated with some scale-related tasks when they learn more about the objects with the similar scale, we propose an easy but efficient scale transformation approach for tiny person detection by keeping the scale consistency between the TinyPerson and the extra dataset. 

\par\setlength\parindent{1em}For dataset $X$, we define the probability density function of objects' size $s$ in $X$ as $P_{size}(s;X)$. Then we define a scale transform $T$, which is used to transform the probability distribution of objects' size in extra dataset $E$ to that in the targeted dataset $D$ (TinyPerson), given in Eq.(3):
\par\setlength\parindent{5em} 
\begin{small}\begin{equation}P_{size}(s;T(E)) 
\approx P_{size}(s;D).
\end{equation}\end{small}

\par\setlength\parindent{1em}In this paper, without losing generality, MS COCO is used as extra dataset, and Scale Match is used for the scale transformation $T$. 

\subsection{Scale Match}
\noindent$G_{ij}=(x_{ij}, y_{ij}, w_{ij}, h_{ij})$ represents j-th object in image $I_i$ of dataset $E$. The Scale Match approach can be simply described as three steps:
\begin{enumerate}
\item  $Sample \  \hat{s} \sim P_{size}(s;D)$;
\item  $Calculate\ scale\ ratio\ c=\frac{\hat{s}}{AS(G_{ij})}$;
\item  $Resize\ object\ with\ scale\ ratio\ c\ , then\ \hat{G_{ij}} \gets (x_{ij} * c, y_{ij} * c, w_{ij} * c, h_{ij} * c)$;
\end{enumerate}
where $\hat{G_{ij}}$ is the result after Scale Match.  Scale Match will be applied to all objects in $E$ to get $T(E)$, when there are a large number of targets in $E$, $P_{size}(s;T(E))$ will be close to $P_{size}(s;D)$. Details of Scale Match algorithm are shown in Algorithm \ref{alg:ScaleMatch}.

\begin{algorithm}
\small
\caption{Scale Match (SM) for Detection}
\label{alg:ScaleMatch}
\textbf{INPUT:} $D_{train}$ (train set of $D$)
\newline
\textbf{INPUT:} $K$ (integer, number of bin in histogram which use to estimate $P_{size}(s;D_{train})$)
\newline
\textbf{INPUT:} $E$ (extra labeled dataset)
\newline
\textbf{OUTPUT:} $\hat{E}$ (note as $T(E)$ before.)
\newline
\textbf{NOTE:} $H$ is the histogram for estimating $P_{size}(s;D_{train})$; $R$ is the size's range of each histogram bin; $I_i$ is $i$-th image in dataset $E$; $G_i$ represents all ground-truth boxes set in $I_i$; $ScaleImage$ is a function to resize image and gorund-truth boxes with a given scale.
\newline
\begin{algorithmic}[1]
    \State $\hat{E} \gets \emptyset$
    \State // to obtain approximate $P_{size}(s;D_{train})$. 
    
    \State $(H\ , Sizes) \gets RectifiedHistogram(D_{train}\ , K)$

    \For{$(I_i, G_i) \ \ in \ \ E$}
        \State // calculate mean size of box in $G_i$
        \State $s \gets Mean((G_i))$              
        \State // sample a bin index of $H$ by the probability value
        \State sample $k \sim H$
        \State sample $\hat{s} \sim Uniform(R[k]^{-}, R[k]^{+})$
        \State $c \gets \hat{s}/s$
        \State $\hat{I_i},\hat{G_i} \gets ScaleImage(I_i,  G_i, c)$
        \State $\hat{E} \gets \hat{E} \cup {(\hat{I_i}, \hat{G_i})}$
    \EndFor
\end{algorithmic}
\end{algorithm}

\noindent \textbf{Estimate $\bm{P_{size}(s;D)}$:} In Scale Match, we first estimate $P_{size}(s;D)$, following a basic assumption in machine learning: the distribution of randomly sampled training dataset is close to actual distribution. Therefore, the training set $P_{size}(s;D_{train})$ is used to approximate $P_{size}(s;D)$. 

\noindent \textbf{Rectified Histogram:} The discrete histogram $(H, R)$ is used to approximate $P_{size}(s;D_{train})$ for calculation, $R[k]^-$ and $R[k]^+$ are size boundary of $k$-th bin in histogram, $K$ is the number of bins in histogram, $N$ is the number of objects in $D_{train}$, $G{ij}(D_{train}) $is j-th object in i-th image of dataset $D_{train}$, and $H[k]$ is probability of $k$-th bin given in Eq (4):

\begin{small}\begin{equation}H[k] = \frac{|\{G_{ij}(D_{train})|R[k]^- \le AS(G_{ij}(D_{train})) < R[k]^+\}|}{N}.\end{equation}\end{small}

However, the long tail of dataset distribution (shown in Figure \ref{fig:scale match}) makes histogram fitting inefficient, which means that  many bins' probability is close to 0. Therefore, a more efficient rectified histogram (as show in Algorithm \ref{alg:RectifiedHistogram}) is proposed.
And $SR$ (sparse rate), calculating how many bins' probability are close to 0 in all bins, is defined as the measure of $H$'s fitting effectiveness:

\begin{small}
\begin{equation}SR = \frac{|\{k|\ H[k] \le 1/(\alpha*K)\ and\ k=1,2...,K|}{K}.\end{equation}
\end{small}

\noindent where $K$ is defined as the bin number of the $H$ and is set to 100, $\alpha$ is set to 10 for $SR$, and $1/(\alpha*K)$ is used as a threshold. With rectified histogram, $SR$ is down to 0.33 from 0.67 for TinyPerson. The rectified histogram $H$ pays less attention on long tail part which has less contribution to distribution. 

\noindent \textbf{Image-level scaling:} For all objects in extra dataset $E$, we need sample a $\hat{s}$ respect to $P_{size}(s; D_{train})$ and resize the object to $\hat{s}$. In this paper, instead of resizing the object, we resize the image which hold the object to make the object's size reach $\hat{s}$. Due to only resizing these objects will destroy the image structure. However there are maybe more than one object with different size in one image. 
We thus sample one $\hat{s}$ per image and guarantees the mean size of objects in this image to $\hat{s}$. 

\noindent \textbf{Sample \bm{$\hat{s}$}:} We firstly sample a bin's index respect to probability of $H$, and secondly sample $\hat{s}$ respect to a uniform probability distribution with min and max size equal to $R[k]^-$ and $R[k]^+$. The first step ensures that the distribution of $\hat{s}$ is close to that of $P_{size}(s;D_{train)}$. For the second step, a uniform sampling algorithm is used.


\begin{algorithm}
\small
\caption{Rectified Histogram}
\label{alg:RectifiedHistogram}
\textbf{INPUT:} $D_{train}$ (train dataset of $D$)
\newline
\textbf{INPUT:} $K$ (integer, $K>2$)
\newline
\textbf{OUTPUT:} $H$ (probability of each bin in the histogram for estimating $P_{size}(s;D_{train})$)

\textbf{OUTPUT:} $R$ (size's range of each bin in histogram)
\newline
\textbf{NOTE:} $N$ (the number of objects in dataset $D$); $G_{ij}(D_{train})$ is j-th object in i-th image of dataset $D_{train}$.
\newline
\begin{algorithmic}[1]
    \State Array $R[K], H[K]$
    \State // collect all boxes' size in $D_{train}$ as $S_{all}$
    \State $S_{all} \gets (..., AS(G_{ij}(D_{train})), ...)$
    
    \State // ascending sort
    \State $S_{sort} \gets sorted(S_{all})$  
    \State
    \State // rectify part to solve long tail
    \State $\color{black}{p \gets \frac{1}{K}}$
    \State $\color{black}{N \gets |S_{sort}|}$
    \State // first $tail$ small boxes' size are merge to first bin
    \State $\color{black}{tail \gets \ ceil(N*p)}$
    \State $\color{black}{R[1]^{-} \gets min(S_{sort})}$
    \State $\color{black}{R[1]^+ \gets S_{sort}[tail+1]}$
    \State $\color{black}{H[1] \gets \frac{tail}{N}}$
    \State // last $tail$ big boxes' size are merge to last bin
    \State $\color{black}{R[K]^- \gets S_{sort}[N-tail]}$
    \State $\color{black}{R[K]^+ \gets max(S_{sort})}$
    \State $\color{black}{H[K] \gets \frac{tail}{N}}$
    \State
    \State $S_{middle} \gets S_{sort}[tail+1:N-tail]$
    \State // calculate histogram with uniform size step and have $K-2$ bins for $S_{middle}$ to get $H[2], H[3], ..., H[K-1]$ and $R[2], R[3], ..., R[K-1]$.
    \State $d \gets \frac{max(S_{middle}\ )-min(S_{middle}\ )}{K-2}$
    \For{$k\ in\ 2,3,...,K-1$}
    \State $R[k]^- \gets min(S_{middle}) + (k -2) * d$ 
    \State $R[k]^+ \gets min(S_{middle}) + (k-1)*d$
    \State $H[k] = \frac{|\{G_{ij}(D_{train})|R[k]^- \le AS(G_{ij}(D_{train})) < R[k]^+\}|}{N}$
    \EndFor
   
\end{algorithmic}
\end{algorithm}

\begin{figure}
    \begin{center}
        \includegraphics[width=7cm]{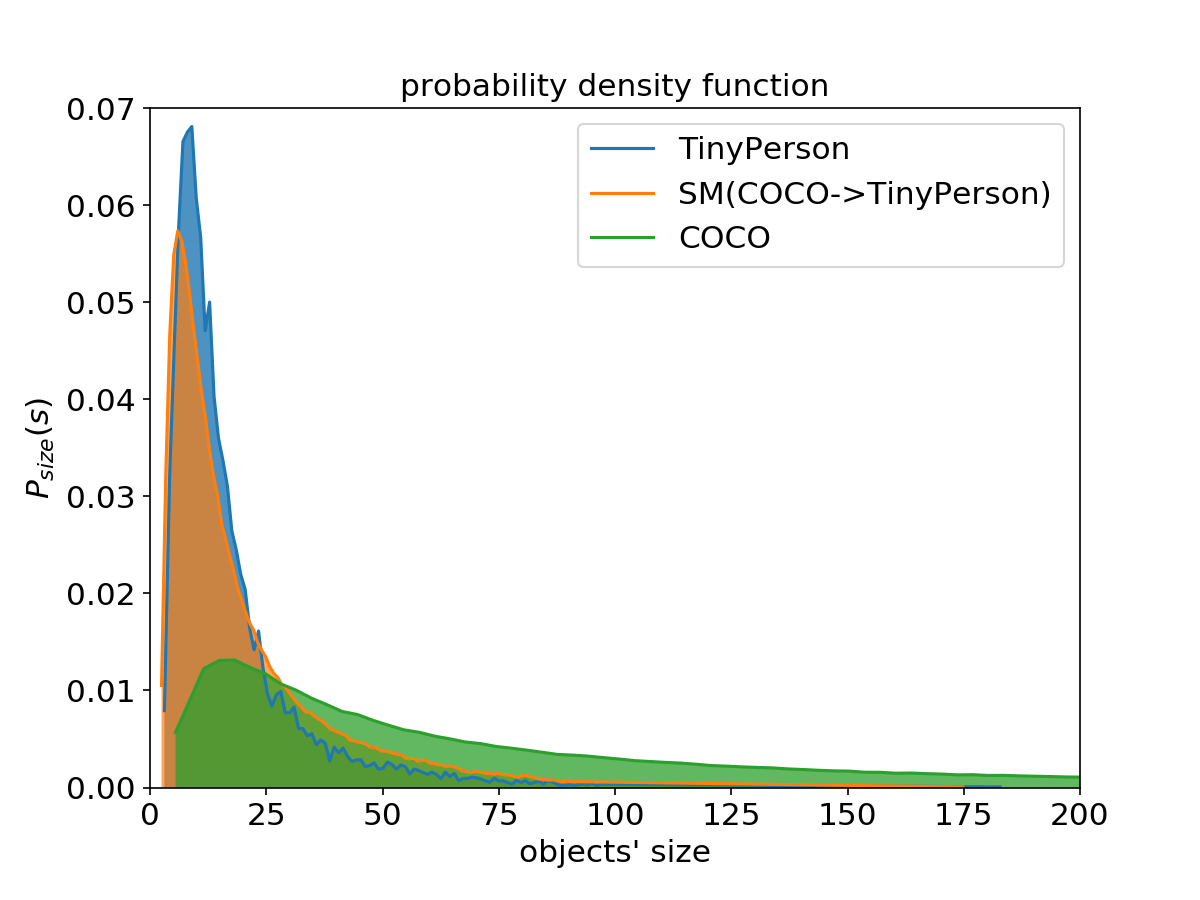}
    \end{center}
    \caption{$P_{size}(s;X)$ of COCO, TinyPerson and COCO after Scale Match to TinyPerson, for better view, we limit the max object's size to 200 instead of 500.}
    \label{fig:SM(coco->TinyPerson)}
\end{figure}

\subsection{Monotone Scale Match (MSM) for Detection}

\begin{figure}
    \begin{center}
        \includegraphics[width=8cm]{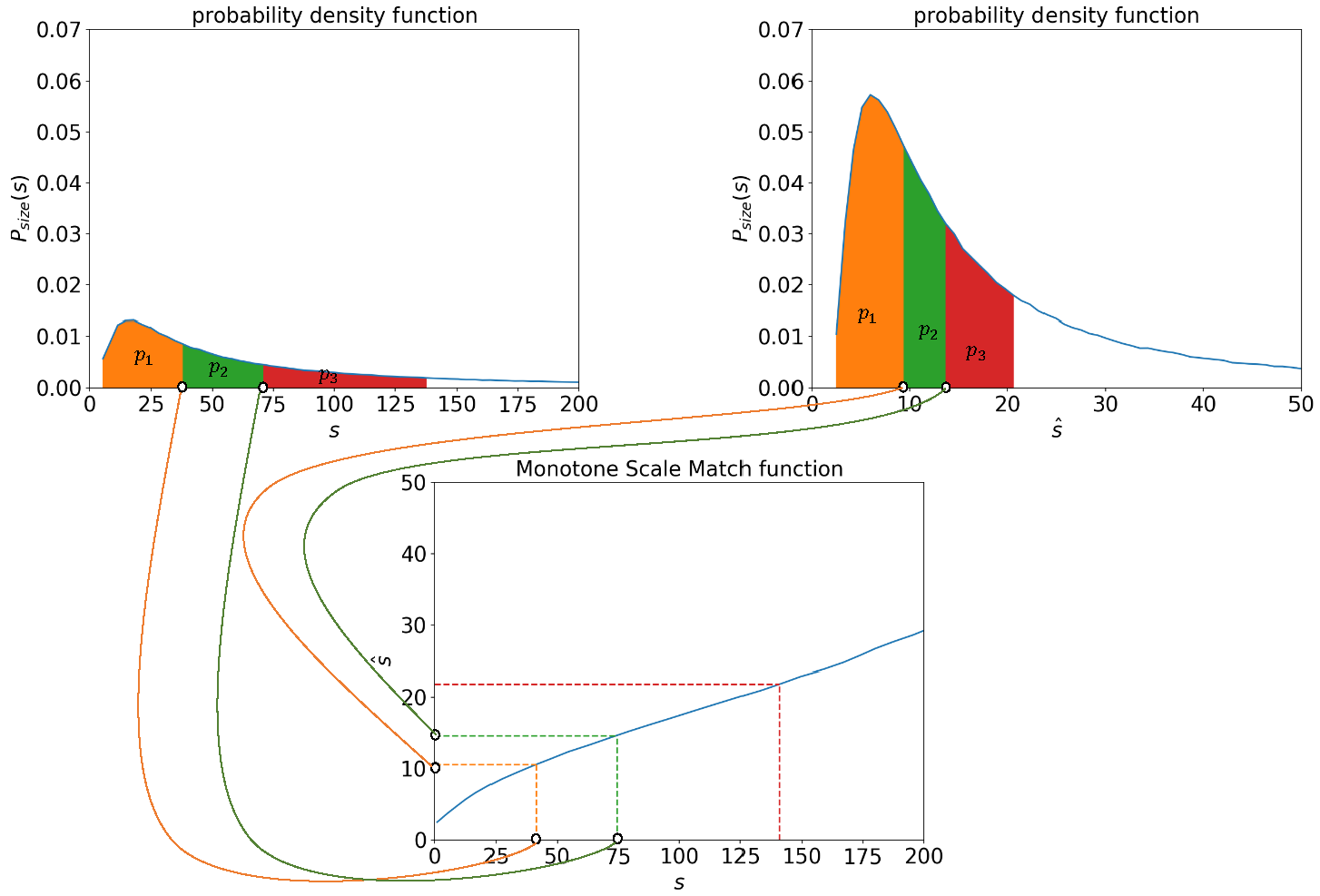}
    \end{center}
    \caption{The flowchart of Monotone Scale Match, mapping the object's size $s$ in $E$ to $\hat{s}$ in $\hat{E}$ with a monotone function.}
    \label{fig:MSM}
\end{figure}

\noindent Scale Match can transform the distribution of size to task-specified dataset, as shown in Figure \ref{fig:SM(coco->TinyPerson)}. Nevertheless, Scale Match may make the original size out of order: a very small object could sample a very big size and vice versa. The Monotone Scale Match, which can keep the monotonicity of size, is further proposed for scale transformation.

It is known that the histogram Equalization and Matching algorithms for image enhancement keep the monotonic changes of pixel values. We follow this idea monotonically change the size, as shown in Figure \ref{fig:MSM}. Mapping object's size $s$ in dataset $E$ to $\hat{s}$ with a monotone function $f$, makes the distribution of $\hat{s}$ same as $P_{size}(\hat{s}, D_{train})$. For any $s_{0} \in [min(s), max(s)]$, it is calculated as:

\begin{small}
\begin{equation}
\int_{min(s)}^{s_{0}}P_{size}(s;E)ds = \int_{f(min(s))}^{f(s_{0})}P_{size}(\hat{s};D_{train})d\hat{s}.
\end{equation}\end{small}

\noindent where $min(s)$ and $max(s)$ represent the minimum and maximum size of objects in $E$, respectively.

\section{Experiments}
\subsection{Experiments Setting}
\noindent \textbf{Ignore region:} In TinyPerson, we must handle ignore regions in training set. Since the ignore region is always a group of persons (not a single person) or something else which can neither be treated as foreground (positive sample) nor background (negative sample). There are two ways for processing the ignore regions while training: 1) Replace the ignore region with mean value of the images in training set; 2) Do not back-propagate the gradient which comes from ignore region. In this paper, we just simply adopt the first way for ignore regions. 
\newline
\textbf{Image cutting:} Most of images in TinyPerson are with large size, results in the GPU out of memory. Therefore, we cut the origin images into some sub-images with overlapping during training and test. Then the NMS strategy is used to merge all results of the sub-images in one same image for evaluation. Although the image cutting can make better use of GPU resources, there are two flaws:1) For FPN, pure background images (no object in this image) will not be used for training. Due to image cutting, many sub-images will become the pure background images, which are not well utilized; 2) In some conditions, NMS can not merge the results in overlapping regions well. 
\newline
\textbf{Training detail:} The codes are based on facebook maskrcnn-benchmark. We choose ResNet50 as backbone. If no specified, Faster RCNN-FPN are chose as detector. Training 12 epochs, and base learning rate is set to 0.01, decay 0.1 after 6 epochs and 10 epochs. We train and evaluate on two 2080Ti GPUs. Anchor size is set to (8.31, 12.5, 18.55, 30.23, 60.41), aspect ratio is set to (0.5, 1.3, 2) by clustering. Since some images are with dense objects in TinyPerson, DETECTIONS\_PER\_IMG (the max number of detector’s output result boxes per image) is set to 200.
\newline
\textbf{Data Augmentation:} Only flip horizontal is adopted to augment the data for training. Different from other FPN based detectors, which resize all images to the same size, we use the origin image/sub-image size without any zooming. 

\subsection{Baseline for TinyPerson Detection}

\setlength{\tabcolsep}{7pt}\begin{table*}
    \centering
    \small
    \begin{tabular}{|l|c|c|c|c|c||c|c|}
        \hline
        detector & $MR^{tiny1}_{50}$ & $MR^{tiny2}_{50}$ &  $MR^{tiny3}_{50}$ & $MR^{tiny}_{50}$ & $MR^{small}_{50}$ & $MR^{tiny}_{25}$ & $MR^{tiny}_{75}$\\
        \hline \hline
        FCOS \cite{tian2019fcos}						& 99.10& 96.39& 91.31& 96.12& 84.14& 89.56&99.56\\
        RetinaNet \cite{lin2017focal}					& 95.05& 88.34&	86.04& 92.40& 81.75& 81.56& 99.11\\
        DSFD \cite{li2019dsfd}                        & 96.41& 88.02& 86.84& 93.47& 78.72& 78.02& 99.48\\
        Adaptive RetinaNet          & 89.48& 82.29& 82.40& 89.19& 74.29& 77.83& 98.63 \\
        Adaptive FreeAnchor \cite{zhang2019freeanchor}         & 90.26& 82.01& 81.74& 88.97& 73.67& 77.62& 98.7 \\
        Faster RCNN-FPN \cite{lin2017feature}		        & 88.40& 81.99& 80.17& 87.78& 71.31& 77.35& 98.40 \\
        \hline
    \end{tabular}
    \caption{Comparisons of $MR$s on TinyPerson.}
\label{tab:baseline mr}
\end{table*}

\setlength{\tabcolsep}{9pt}\begin{table*}
    \centering
    \small
    \begin{tabular}{|l|c|c|c|c|c||c|c|}
        \hline
        detector & $AP^{tiny1}_{50}$ & $AP^{tiny2}_{50}$ &  $AP^{tiny3}_{50}$ & $AP^{tiny}_{50}$ & $AP^{small}_{50}$& $AP^{tiny}_{25}$& $AP^{tiny}_{75}$\\
        \hline \hline
        FCOS \cite{tian2019fcos}						& 3.39 & 12.39& 29.25& 16.9& 35.75& 40.49& 1.45\\
        RetinaNet \cite{lin2017focal}					& 11.47& 36.36&	43.32& 30.82& 43.38& 57.33& 2.64\\
        DSFD \cite{li2019dsfd}                        & 14.00& 35.12& 46.30& 31.15& 51.64& 59.58& 1.99\\
        Adaptive RetinaNet          & 25.49& 47.89& 51.28& 41.25& 53.64& 63.94& 4.22\\
        Adaptive FreeAnchor \cite{zhang2019freeanchor}         & 24.92& 48.01& 51.23& 41.36& 53.36& 63.73& 4.0\\
        Faster RCNN-FPN	\cite{lin2017feature}	        & 29.21& 48.26& 53.48& 43.55& 56.69& 64.07& 5.35\\
        \hline
    \end{tabular}
    \caption{Comparisons of $AP$s on TinyPerson.}
\label{tab:baseline ap}
\end{table*}

\noindent For TinyPerson, the RetinaNet\cite{lin2017focal}, FCOS\cite{tian2019fcos}, Faster RCNN-FPN, which are the representatives of one stage anchor base detector, anchor free detector and two stage anchor base detector respectively, are selected for experimental comparisons. To guarantee the convergence, we use half learning rate of Faster RCNN-FPN for RetinaNet and quarter for FCOS. For adaptive FreeAnchor\cite{zhang2019freeanchor}, we use same learning rate and backbone setting of Adaptive RetinaNet, and others are keep same as FreeAnchor's default setting. 

In Figure \ref{fig:motivation}, WIDER Face holds a similar absolute scale distribution to TinyPerson. Therefore, the state-of-the-art of DSFD detector \cite{li2019dsfd}, which is specified for tiny face detection, has been included for comparison on TinyPerson.

\noindent\textbf{Poor localization:} As shown in Table \ref{tab:baseline mr} and Table \ref{tab:baseline ap}, the performance drops significantly while IOU threshold changes from 0.25 to 0.75. It's hard to have high location precision in TinyPerson due to the tiny objects' absolute and relative size.
\newline
\noindent\textbf{Spatial information:} Due to the size of the tiny object, spatial information maybe more important than deeper network model. Therefore, we use P2, P3, P4, P5, P6 of FPN instead of P3, P4, P5, P6, P7 for RetinaNet, which is similar to Faster RCNN-FPN. We named the adjusted version as Adaptive RetinaNet. It achieves better performance (10.43\%\ improvement of $AP^{tiny}_{50}$) than the RetinaNet. 
\newline
\noindent\textbf{Best detector:} With MS COCO, RetinaNet and FreeAnchor achieves better performance than Faster RCNN-FPN. One stage detector can also go beyond two stage detector if sample imbalance is well solved \cite{lin2017focal}. The anchor-free based detector FCOS achieves the better performance compared with RetinaNet and Faster RCNN-FPN. However, when objects' size become tiny such as objects in TinyPerson, the performance of all detectors drop a lot. And the RetinaNet and FCOS performs worse, as shown in Table \ref{tab:baseline mr} and Table \ref{tab:baseline ap}. For tiny objects, two stage detector shows advantages over one stage detector. 
Li et al. \cite{li2019dsfd} proposed DSFD for face detection, which is one of the SOTA face detectors with code available. But it obtained poor performance on TinyPerson, due to the great difference between relative scale and aspect ratio, which also further demonstrates the great chanllange of the proposed TinyPerson. With performance comparison, Faster RCNN-FPN is chosen as the baseline of experiment and the benchmark.

\subsection{Analysis of Scale Match}
\noindent\textbf{TinyPerson.} In general, for detection, pretrain on MS COCO often gets better performance than pretrain on ImageNet, although the ImageNet holds more data. However, detector pre-trained on MS COCO improves very limited in TinyPerson, since the object size of MS COCO is quite different from that of TinyPerson. Then, we obtain a new dataset, COCO100, by setting the shorter edge of each image to 100 and keeping the height-width ratio unchanged. The mean of objects' size in COCO100 almost equals to that of TinyPerson. However, the detector pre-trained on COCO100 performs even worse, shown in Table \ref{tab:scale matching}. The transformation of the mean of objects' size to that in TinyPerson is inefficient. Then we construct SM COCO by transforming the whole distribution of MS COCO to that of TinyPerson based on Scale Match. With detector pre-trained on SM COCO, we obtain 3.22\%\ improvement of $AP_{50}^{tiny}$, Table \ref{tab:scale matching}. Finally we construct MSM COCO using Monotone Scale Match for transformation of MS COCO. With MSM COCO as the pre-trained dataset, the performance further improves to 47.29\%\ of $AP_{50}^{tiny}$, Table \ref{tab:scale matching}.

\noindent\textbf{Tiny Citypersons.} To further validate the effectiveness of the proposed Scale Match on other datasets, we conducted experiments on Tiny Citypersons and obtained similar performance gain, Table \ref{tab:scale matching tiny cityperson}.

\begin{table}[H]
    \centering
    \small
    \begin{tabular}{|c|c|c|c}
        \hline
        pretrain dataset & $MR^{tiny}_{50}$  & $AP^{tiny}_{50}$\\
        \hline\hline
        ImageNet & 87.78 & 43.55 \\
        COCO     & 86.58 & 43.38 \\
        COCO100  & 87.67 & 43.03 \\
        SM COCO  & 86.30 & {46.77} \\
        MSM COCO & \textbf{85.71} & \textbf{47.29} \\
        \hline
    \end{tabular}
    \caption{Comparisons on \textbf{TinyPerson}. COCO100 holds the similar mean of the boxes' size with TinyPerson, SM COCO uses Scale Match on COCO for pre-training, while MSM COCO uses Monotonous Scale Match on COCO for pre-training.}
    \label{tab:scale matching}
\end{table}

\begin{table}[H]
    \centering
    \small
    \begin{tabular}{|c|c|c|c}
        \hline
        pretrain dataset & $MR^{tiny}_{50}$  & $AP^{tiny}_{50}$\\
        \hline\hline
        ImageNet & 75.44 & 19.08 \\
        COCO     & 74.15 & 20.74 \\
        COCO100  & 74.92 & 20.57 \\
        SM COCO  & 73.87 & 21.18 \\
        MSM COCO & \textbf{72.41} & \textbf{21.56} \\
        \hline
    \end{tabular}
    \caption{Comparisons on \textbf{Tiny Citypersons}. COCO100 holds the similar mean of the boxes' size with Tiny Citypersons.}
    \label{tab:scale matching tiny cityperson}
\end{table}

\section{Conclusion}

\noindent In this paper, a new dataset (TinyPerson) is introduced for detecting tiny objects, particularly, tiny persons less than 20 pixels in large-scale images. The extremely small objects raise a grand challenge for existing person detectors. 

We build the baseline for tiny person detection and experimentally find that the scale mismatch could deteriorate the feature representation and the detectors. We thereby proposed an easy but efficient approach, Scale Match, for tiny person detection. Our approach is inspired by the Human Cognition Process, while Scale Match can better utilize the existing annotated data and make the detector more sophisticated. Scale Match is designed as a plug-and-play universal block for object scale processing, which provides a fresh insight for general object detection tasks. 

{\small
\bibliographystyle{ieee}
\bibliography{egbib}
}

\end{document}